\documentclass[runningheads]{llncs}
\usepackage[T1]{fontenc}
\usepackage{graphicx}
\usepackage{hyperref}
\usepackage{color}

\usepackage[table,xcdraw]{xcolor} 
\usepackage{caption}  
\usepackage{cleveref} 
\begin{document}
\title{e-Fold Cross-Validation for Recommender-System Evaluation}
%
%
\author{Moritz Baumgart\inst{1}\orcidID{0009-0007-1322-1450} \and
Lukas Wegmeth\inst{1}\orcidID{0000-0001-8848-9434} \and
Tobias Vente\inst{1}\orcidID{0009-0003-8881-2379} \and
Joeran Beel\inst{1,2}\orcidID{0000-0002-4537-5573}}
\authorrunning{Baumgart et al.}
%
\institute{University of Siegen, Department of Electrical Engineering and Computer Science, Germany \and
Recommender-Systems.com, Siegen, Germany\\
\email{joeran.beel@uni-siegen.de}\\
\url{https://www.recommender-systems.com} }
\maketitle              
\begin{abstract}
To combat the rising energy consumption of recommender systems we implement a novel alternative for k-fold cross validation. This alternative, named e-fold cross validation, aims to minimize the number of folds to achieve a reduction in power usage while keeping the reliability and robustness of the test results high. We tested our method on 5 recommender system algorithms across 6 datasets and compared it with 10-fold cross validation. On average e-fold cross validation only needed 41.5\% of the energy that 10-fold cross validation would need, while it's results only differed by 1.81\%. We conclude that e-fold cross validation is a promising approach that has the potential to be an energy efficient but still reliable alternative to k-fold cross validation.

\keywords{Energy efficiency \and Cross validation \and Sustainability.}
\end{abstract}

\section{Introduction}
\label{sec:intro}
As the recommender systems community moves towards more deep learning based models \cite{Mu2018,Zhang2019}, it also faces the problem that a higher energy consumption can be observed \cite{Gupta2021,Yang2017}. Recent research even suggests that a paper on recommender systems, which uses deep learning algorithms, needs around 42 times more energy than a paper that uses traditional algorithms \cite{Vente2024}. A higher energy consumption leads to more carbon emissions, which are one of the main causes for climate change and all the negative consequences that come with it \cite{Crimmins2016}. Therefore, we should strive towards more sustainable development of recommender systems and make model learning more energy efficient and low-carbon emitting.

To ensure the robustness of test results, k-Fold Cross-Validation (\textit{k-CV}) is a favored approach often used in practice today \cite{Chen2023,Li2023a,Zhu2023}. \textit{k-CV} splits the dataset into $k$ folds and conducts tests on each one of them, while the other $k - 1$ folds are used for training \cite{Refaeilzadeh2009,Yadav2016,Zhang2015}. The issue with this technique is, that it increases the power consumption roughly by the factor $k$. This is especially problematic when the $k$ is chosen arbitrarily, which often seems to be the case \cite{Anguita2012}.

\label{sec:relatedWork}
Recently, the recommender-system community started to investigate "Green Recommender Systems" \cite{Beel2024e}, i.e. methods to minimize environmental impact. Also, we proposed tools to measure energy consumption \cite{Wegmeth2024a} and save energy through downsampling datasets \cite{Arabzadeh2024}. The machine learning community started already some years ago to explore "Green" (Automated) Machine Learning \cite{tornede2023towards}, and explored an early-stopping approach that is similar to ours \cite{bergman2024dontwastetimeearly}.

While there is lots of research that tries to find an optimal $k$, there seems to be no one who chooses $k$ from an energy saving perspective. Marcot and Hanea \cite{Marcot2021} try out different values for $k$. They support the common use of $k = 10$, but they also acknowledge that in some cases $k = 5$ is sufficient. This indicates the potential to save energy by determining when a smaller $k$ is enough. Arlot and Celisse \cite{Arlot2016} support such a range for $k$ and conclude that a value between $5$ and $10$ is optimal with no statistical improvement for larger values. Similarly, Kohavi and John \cite{Kohavi2001} recommend choosing $k = 10$. Anguita et al. \cite{Anguita2012} consider $k$ as a tunable hyperparameter. While this approach produces an statistically robust value for $k$, it only further increases the power consumption needed to tune the value.

To address the relatively high energy consumption of \textit{k-CV}, we recently proposed the idea of ''e-fold cross-validation'' (\textit{e-CV}), which replaces the often arbitrarily chosen $k$ with an intelligently chosen parameter $e$ \cite{Beel2024}. Our intension was that $e$ is chosen as small as possible to maximize energy savings, but large enough to provide robust results. The first results in the general machine learning domain were promising \cite{Mahlich2024}.  

Our proposal \cite{Beel2024} sets the goal of our current paper and was the namesake of the algorithm described in the following sections. We want to create and evaluate the first possible implementation of \textit{e-CV} with a focus on recommender systems. Our key idea is, that we halt the folding process early, once a certain confidence in the test results is reached.

\section{Methodology}
\label{sec:methodology}
We trained 5 algorithms on 6 datasets using 10-CV and tested them using a top-n prediction task with NDCG@10 as a metric. By incrementally providing \textit{e-CV} with the obtained scores, we then \textit{simulate} how it would operate. Since the order in which the folds happen is arbitrary, we looked at 5000 permutations and did all further evaluations for each permutation individually. Considering all possible permutations was unfeasible, so we only looked at a random subset of 5000. As discussed before, a value of $k$ between 5 and 10 is typical, so we compared our \textit{e-CV} implementation with 10-CV as ground truth. The comparison then works as follows: We calculated the percentage difference $d(x, y) = \frac{|x - y|}{(x + y) \div 2} \cdot 100$ between the final \textit{e-CV} score and the 10-CV score and we record the stopping point calculated by \textit{e-CV}. This enables us to compare how much the results differ for how much energy saving. Lastly, we also ranked the algorithms using the \textit{e-CV} scores and using the 10-CV scores to verify ranking consistency between both methods.

To cover a wide variety of algorithm types, the following algorithms were chosen: From LensKit\cite{Ekstrand2020} we chose \textit{ImplicitMF} as a matrix-factorization method, and from RecBole\cite{Zhao2020} we chose \textit{ItemKNN}\cite{Deshpande2004} as a ''traditional'' algorithm, \textit{MultiVAE}\cite{Liang2018} as an autoencoder, \textit{NeuMF}\cite{He2017} as a deep learning model and \textit{Popularity based (Pop)} as a simple baseline.

\begin{figure}
    \begin{minipage}{0.55\textwidth}
        \resizebox{\textwidth}{!}{
            \begin{tabular}{|l|l|l|l|l|}
            \hline
            \multicolumn{1}{|c|}{\textbf{Dataset   Name}}                                         & \multicolumn{1}{c|}{\textbf{\#Users}} & \multicolumn{1}{c|}{\textbf{\#Items}} & \multicolumn{1}{c|}{\textbf{\#Interactions}} & \multicolumn{1}{c|}{\textbf{Density}} \\ \hline
            \rowcolor[HTML]{D9D9D9} 
            \begin{tabular}[c]{@{}l@{}}Amazon2014\\      Cell-Phones-And-Accessories\cite{He2016}\end{tabular} & 27879                                 & 10429                                 & 194439                                       & 0,0669                                \\ \hline
            \begin{tabular}[c]{@{}l@{}}Amazon2014\\      Apps-For-Android\cite{He2016}\end{tabular}            & 87271                                 & 13209                                 & 752937                                       & 0,0653                                \\ \hline
            \rowcolor[HTML]{D9D9D9} 
            \begin{tabular}[c]{@{}l@{}}Amazon2014\\      Amazon-Instant-Video\cite{He2016}\end{tabular}        & 5130                                  & 1685                                  & 37126                                        & 0,4295                                \\ \hline
            Hetrec-LastFM\cite{Cantador2011}                                                                         & 1090                                  & 3646                                  & 52551                                        & 1,3223                                \\ \hline
            \rowcolor[HTML]{D9D9D9} 
            MovieLens-100K\cite{Harper2015}                                                                        & 943                                   & 1349                                  & 99287                                        & 7,8049                                \\ \hline
            MovieLens-1M\cite{Harper2015}                                                                        & 6040                                  & 3416                                  & 999611                                       & 4,8448                                \\ \hline
            \end{tabular}
        }
        \captionof{table}{Overview of used dataset after preprocessing}
        \label{tab:ds_stats}
    \end{minipage}
    \hfill
    \begin{minipage}{0.4\textwidth}
        \includegraphics[width=\textwidth]{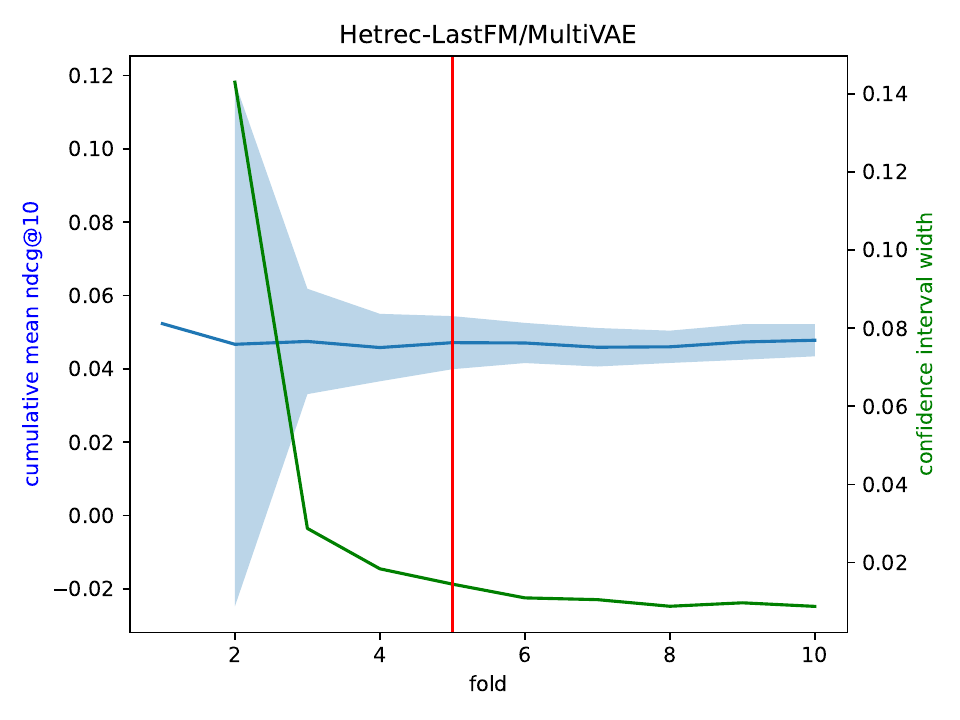}
        \caption{Exemplary \textit{e-CV} run}
        \label{fig:stop_example}
    \end{minipage}
\end{figure}

We chose the following datasets: Three subsets from the Amazon2014 dataset \cite{He2016} and two differently sized MovieLens\cite{Harper2015} datasets, which contain product reviews from Amazon and movie ratings from the MovieLens website respectively, as well as the HetRec 2011 Last.FM dataset\cite{Cantador2011}, which contains artist listening records from users of the Last.fm online music system. For preprocessing we orientate ourselves on other papers and apply 5-core pruning \cite{Kang2018,Sun2019,Vente2024} and convert explicit feedback to implicit feedback by counting the presence of a rating as positive feedback \cite{Chen2023,Ma2023,Vente2024}. An overview of the datasets after this preprocessing is shown in \cref{tab:ds_stats}. To prepare the datasets for 10-CV they were each split into 10 partitions, for each one it was ensured that each user is equally represented. If a user had less than 10 interactions, these were assigned to partitions randomly.

Our proposed implementation calculates the mean of all scores it has so far, as well as the confidence interval (CI) of that mean. It then uses a criterion on the CI width to stop folding. Let $C = \{c_1,...,c_n\}$ be the set of CI widths then we stop folding if $|c_{n-1} - c_n| \le \frac{\alpha}{c_n}$, with $\alpha$ being a user-selectable parameter which can be used to prioritize energy-saving (large $\alpha$) or accuracy (small $\alpha$). In \cref{fig:stop_example} an example of the algorithm working can be seen. It shows the mean of all scores up to the current fold and the confidence interval in blue on the left axis. The right axis shows the width of the confidence interval in green. The red vertical line is the fold at which the stopping criterion was first met.

\section{Results \& Discussion}
On average, across all datasets/algorithms, \textit{e-CV} differs only by 1.81\% from 10-CV, while it stops already after 4.15 folds, which means that \textit{e-CV} uses only $41.5\%$ of the energy.

\begin{figure}
    \centering
    \includegraphics[width=0.9\textwidth]{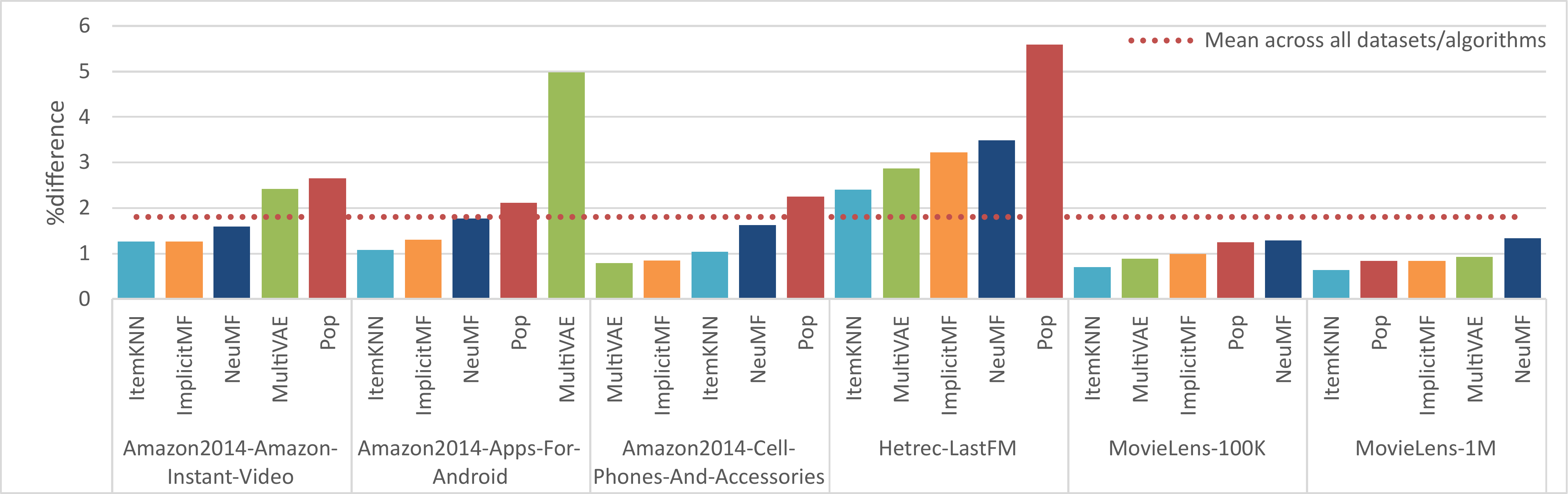}
    \caption{Percentage difference between final \textit{e-CV} score and 10-CV score for each dataset/algorithm, averaged across tested permutations.}
    \label{fig:percent_diff}
\end{figure}

The average percentage difference for each dataset/algorithm can be seen in \cref{fig:percent_diff}. \textit{e-CV} seems to work especially well for the ItemKNN algorithm, while it does not perform as well for MultieVAE and Pop. We can also see that it works very well on the two MovieLens datasets, which both have a higher density than the rest. Interestingly, it did not perform so well on the LastFM dataset, even though this one has a higher density than all the Amazon datasets.

\begin{figure}
    \includegraphics[width=0.9\textwidth]{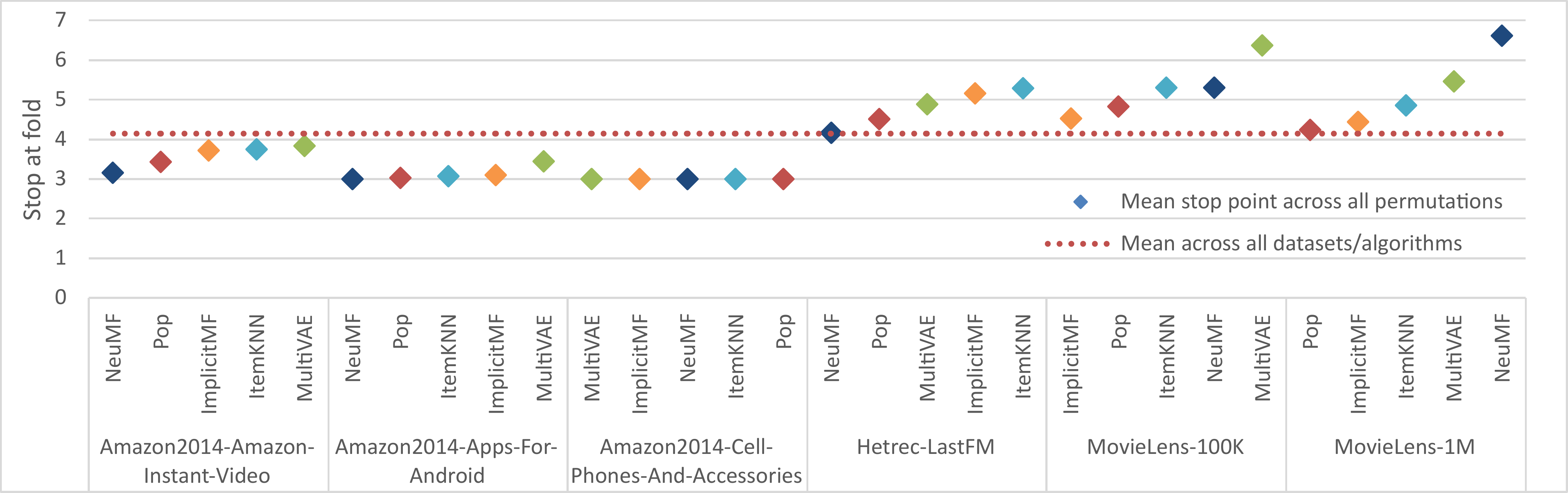}
    \caption{Stopping point determined by \textit{e-CV} for each dataset/algorithm, averaged across tested permutations.}
    \label{fig:stop_points}
\end{figure}
    
\begin{figure}
    \includegraphics[width=\textwidth]{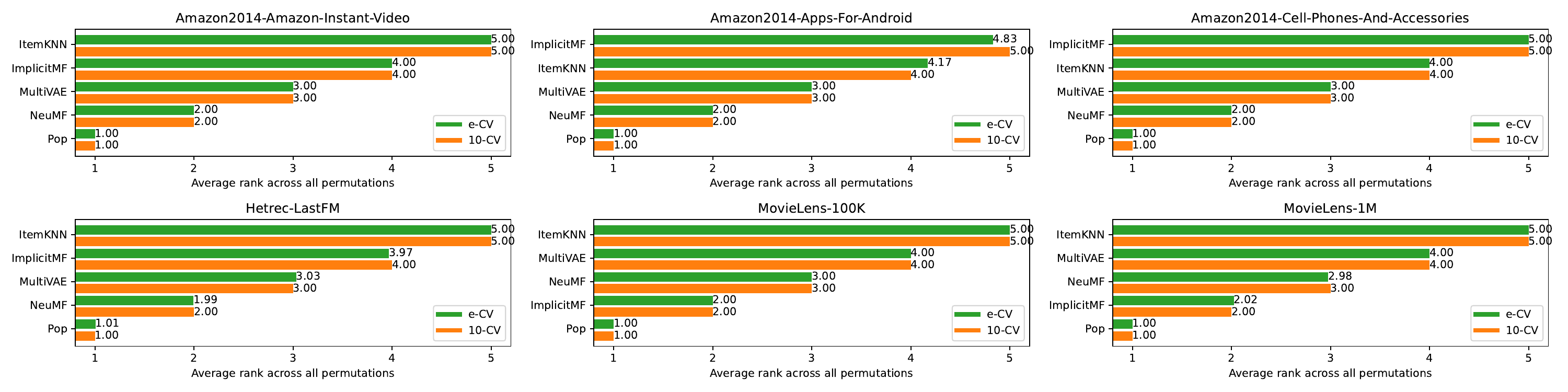}
    \caption{Ranking of the algorithms, averaged across tested permutations.}
    \label{fig:algo_ranking}
\end{figure}

Looking at \cref{fig:stop_points}, we can see at which fold \textit{e-CV} decided to stop the folding process. It is notable that for the MovieLens datasets it stopped at later folds, which could explain the smaller percentage difference for these datasets in \cref{fig:percent_diff}. This, however, does not hold for the LastFM dataset, where both percentage difference and stopping point have higher values, which suggests that \textit{e-CV} did not work well on this dataset in general. For the Amazon datasets and the algorithms ItemKNN, ImplicitMF, and NeuMF \textit{e-CV} seem to have made the right decisions since the percentage difference is low without an abnormal high stopping point.

In \cref{fig:algo_ranking}, we can see that the average ranking of the algorithms for a dataset stayed consistent with \textit{e-CV}. A slight deviation can be observed only for a few selected datasets/algorithms.

All in all, we can say that the idea of \textit{e-CV} seems to have a lot of potential. Especially, since we could achieve these promising results with a rather simple implementation. We think that more sophisticated implementations that consider more factors are likely to perform even better and fix the weaknesses of our current implementation.

\bibliographystyle{splncs04}
\bibliography{refs}

\begin{thebibliography}{10}
\providecommand{\url}[1]{\texttt{#1}}
\providecommand{\urlprefix}{URL }
\providecommand{\doi}[1]{https://doi.org/#1}

\bibitem{Anguita2012}
Anguita, D., Ghelardoni, L., Ghio, A., Oneto, L., Ridella, S.: The 'k' in k-fold cross validation. In: 20th European Symposium on Artificial Neural Networks, {ESANN} 2012, Bruges, Belgium, April 25-27, 2012 (2012), \url{https://www.esann.org/sites/default/files/proceedings/legacy/es2012-62.pdf}

\bibitem{Arabzadeh2024}
Arabzadeh, A., Vente, T., Beel, J.: Green recommender systems: Optimizing dataset size for energy-efficient algorithm performance. In: International Workshop on Recommender Systems for Sustainability and Social Good (RecSoGood) at the 18th ACM Conference on Recommender Systems (ACM RecSys) (2024), \url{https://isg.beel.org/pubs/2024-Green-RecSys-Dataset-Sampling-Ardalan.pdf}

\bibitem{Arlot2016}
Arlot, S., Celisse, A.: A survey of cross-validation procedures for model selection. Statistics Surveys  \textbf{4}(none),  1--50 (2016). \doi{10.1214/09-ss054}

\bibitem{Beel2024e}
Beel, J., Said, A., Vente, T., Wegmeth, L.: Green recommender systems – a call for attention. Recommender-Systems.com Blog  (2024). \doi{10.31219/osf.io/5ru2g}, \url{https://isg.beel.org/pubs/2024_Green_Recommender_Systems-A_Call_for_Attention.pdf}

\bibitem{Beel2024}
Beel, J., Wegmeth, L., Vente, T.: e-fold cross-validation: A computing and energy-efficient alternative to k-fold cross-validation with adaptive folds  (Jun 2024). \doi{10.31219/osf.io/exw3j}

\bibitem{bergman2024dontwastetimeearly}
Bergman, E., Purucker, L., Hutter, F.: Don't waste your time: Early stopping cross-validation (2024), \url{https://arxiv.org/abs/2405.03389}

\bibitem{Cantador2011}
Cantador, I., Brusilovsky, P., Kuflik, T.: Second workshop on information heterogeneity and fusion in recommender systems (hetrec2011). In: Proceedings of the fifth ACM conference on Recommender systems. RecSys ’11, ACM (Oct 2011). \doi{10.1145/2043932.2044016}

\bibitem{Chen2023}
Chen, H., Li, X., Lai, V., Yeh, C.C.M., Fan, Y., Zheng, Y., Das, M., Yang, H.: Adversarial collaborative filtering for free. In: Proceedings of the 17th ACM Conference on Recommender Systems. RecSys ’23, ACM (Sep 2023). \doi{10.1145/3604915.3608771}

\bibitem{Crimmins2016}
Crimmins, A., Balbus, J., Gamble, J., Beard, C., Bell, J., Dodgen, D., Eisen, R., Fann, N., Hawkins, M., Herring, S., Jantarasami, L., Mills, D., Saha, S., Sarofim, M., Trtanj, J., Ziska, L.: The Impacts of Climate Change on Human Health in the United States: A Scientific Assessment (2016). \doi{10.7930/j0r49nqx}

\bibitem{Deshpande2004}
Deshpande, M., Karypis, G.: Item-based top-nrecommendation algorithms. ACM Transactions on Information Systems  \textbf{22}(1),  143--177 (Jan 2004). \doi{10.1145/963770.963776}

\bibitem{Ekstrand2020}
Ekstrand, M.D.: Lenskit for python: Next-generation software for recommender systems experiments. In: Proceedings of the 29th ACM International Conference on Information \& Knowledge Management. CIKM ’20, ACM (Oct 2020). \doi{10.1145/3340531.3412778}

\bibitem{Gupta2021}
Gupta, U., Kim, Y.G., Lee, S., Tse, J., Lee, H.H.S., Wei, G.Y., Brooks, D., Wu, C.J.: Chasing carbon: The elusive environmental footprint of computing. In: 2021 IEEE International Symposium on High-Performance Computer Architecture (HPCA). IEEE (Feb 2021). \doi{10.1109/hpca51647.2021.00076}

\bibitem{Harper2015}
Harper, F.M., Konstan, J.A.: The movielens datasets: History and context. ACM Transactions on Interactive Intelligent Systems  \textbf{5}(4),  1--19 (Dec 2015). \doi{10.1145/2827872}

\bibitem{He2016}
He, R., McAuley, J.: Ups and downs: Modeling the visual evolution of fashion trends with one-class collaborative filtering  (Apr 2016). \doi{10.1145/2872427.2883037}

\bibitem{He2017}
He, X., Liao, L., Zhang, H., Nie, L., Hu, X., Chua, T.S.: Neural collaborative filtering. In: Proceedings of the 26th International Conference on World Wide Web. WWW ’17, International World Wide Web Conferences Steering Committee (Apr 2017). \doi{10.1145/3038912.3052569}

\bibitem{Kang2018}
Kang, W.C., McAuley, J.: Self-attentive sequential recommendation  (Aug 2018). \doi{10.48550/ARXIV.1808.09781}

\bibitem{Kohavi2001}
Kohavi, R., John, G.: The Wrapper Approach, vol.~14, pp. 33--50. Springer US (2001). \doi{10.1007/978-1-4615-5725-8_3}, \url{https://www.researchgate.net/publication/2352264_A_Study_of_Cross-Validation_and_Bootstrap_for_Accuracy_Estimation_and_Model_Selection}

\bibitem{Li2023a}
Li, Y., Tang, X., Chen, B., Huang, Y., Tang, R., Li, Z.: Autoopt: Automatic hyperparameter scheduling and optimization for deep click-through rate prediction. In: Proceedings of the 17th ACM Conference on Recommender Systems. RecSys ’23, ACM (Sep 2023). \doi{10.1145/3604915.3608800}

\bibitem{Liang2018}
Liang, D., Krishnan, R.G., Hoffman, M.D., Jebara, T.: Variational autoencoders for collaborative filtering. In: Proceedings of the 2018 World Wide Web Conference on World Wide Web - WWW ’18. pp. 689--698. WWW ’18, ACM Press (2018). \doi{10.1145/3178876.3186150}

\bibitem{Ma2023}
Ma, H., Xie, R., Meng, L., Chen, X., Zhang, X., Lin, L., Zhou, J.: Exploring false hard negative sample in cross-domain recommendation. In: Proceedings of the 17th ACM Conference on Recommender Systems. RecSys ’23, ACM (Sep 2023). \doi{10.1145/3604915.3608791}

\bibitem{Mahlich2024}
Mahlich, C., Vente, T., Beel, J.: From theory to practice: Implementing and evaluating e-fold cross-validation. In: International Conference on Artificial Intelligence and Machine Learning Research (CAIMLR) (2024), \url{https://isg.beel.org/blog/2024/09/16/e-fold-cross-validation/}

\bibitem{Marcot2021}
Marcot, B.G., Hanea, A.M.: What is an optimal value of k in k-fold cross-validation in discrete bayesian network analysis? Comput. Stat.  \textbf{36}(3),  2009--2031 (2021). \doi{10.1007/S00180-020-00999-9}

\bibitem{Mu2018}
Mu, R.: A survey of recommender systems based on deep learning. IEEE Access  \textbf{6},  69009--69022 (2018). \doi{10.1109/access.2018.2880197}

\bibitem{Refaeilzadeh2009}
Refaeilzadeh, P., Tang, L., Liu, H.: Cross-Validation, pp.~1--7. Springer New York (2009). \doi{10.1007/978-1-4899-7993-3_565-2}

\bibitem{Sun2019}
Sun, F., Liu, J., Wu, J., Pei, C., Lin, X., Ou, W., Jiang, P.: Bert4rec: Sequential recommendation with bidirectional encoder representations from transformer. In: Proceedings of the 28th ACM International Conference on Information and Knowledge Management. CIKM ’19, ACM (Nov 2019). \doi{10.1145/3357384.3357895}

\bibitem{tornede2023towards}
Tornede, T., Tornede, A., Hanselle, J., Mohr, F., Wever, M., H{\"u}llermeier, E.: Towards green automated machine learning: Status quo and future directions. arXiv / Journal of Artificial Intelligence Research  \textbf{77},  427--457 (2021 / 2023)

\bibitem{Vente2024}
Vente, T., Wegmeth, L., Said, A., Beel, J.: From clicks to carbon: The environmental toll of recommender systems. In: Proceedings of the 18th ACM Conference on Recommender Systems. p. 580–590. RecSys '24, Association for Computing Machinery, New York, NY, USA (2024). \doi{10.1145/3640457.3688074}, \url{https://arxiv.org/abs/2408.08203}

\bibitem{Wegmeth2024a}
Wegmeth, L., Vente, T., Said, A., Beel, J.: Emers: Energy meter for recommender systems. In: International Workshop on Recommender Systems for Sustainability and Social Good (RecSoGood) at the 18th ACM Conference on Recommender Systems (ACM RecSys) (2024), \url{https://arxiv.org/pdf/2409.15060}

\bibitem{Yadav2016}
Yadav, S., Shukla, S.: Analysis of k-fold cross-validation over hold-out validation on colossal datasets for quality classification. In: 2016 IEEE 6th International Conference on Advanced Computing (IACC). IEEE (Feb 2016). \doi{10.1109/iacc.2016.25}

\bibitem{Yang2017}
Yang, T.J., Chen, Y.H., Emer, J., Sze, V.: A method to estimate the energy consumption of deep neural networks. In: 2017 51st Asilomar Conference on Signals, Systems, and Computers. IEEE (Oct 2017). \doi{10.1109/acssc.2017.8335698}

\bibitem{Zhang2019}
Zhang, S., Yao, L., Sun, A., Tay, Y.: Deep learning based recommender system: A survey and new perspectives. ACM Computing Surveys  \textbf{52}(1),  1--38 (Feb 2019). \doi{10.1145/3285029}

\bibitem{Zhang2015}
Zhang, Y., Yang, Y.: Cross-validation for selecting a model selection procedure. Journal of Econometrics  \textbf{187}(1),  95--112 (7 2015). \doi{10.1016/j.jeconom.2015.02.006}

\bibitem{Zhao2020}
Zhao, W.X., Mu, S., Hou, Y., Lin, Z., Chen, Y., Pan, X., Li, K., Lu, Y., Wang, H., Tian, C., Min, Y., Feng, Z., Fan, X., Chen, X., Wang, P., Ji, W., Li, Y., Wang, X., Wen, J.R.: Recbole: Towards a unified, comprehensive and efficient framework for recommendation algorithms  (Nov 2020). \doi{10.48550/ARXIV.2011.01731}

\bibitem{Zhu2023}
Zhu, J., Wang, Y., Zhu, F., Sun, Z.: Domain disentanglement with interpolative data augmentation for dual-target cross-domain recommendation. In: Proceedings of the 17th ACM Conference on Recommender Systems. RecSys ’23, ACM (Sep 2023). \doi{10.1145/3604915.3608802}

\end{thebibliography}

\typeout{get arXiv to do 4 passes: Label(s) may have changed. Rerun}

\end{document}